\documentclass[11p]{article}
\usepackage{pdfsync}
\usepackage{graphicx}
\usepackage{amssymb,amsfonts,amsmath,dsfont,amsthm}
\usepackage{verbatim}
\usepackage{varwidth}
\usepackage{algorithm,algpseudocode,lipsum}
\usepackage{caption}
\usepackage{framed}
\usepackage{xcolor}
\usepackage{lipsum}
\usepackage{authblk}

\newcommand{\N}{\mathbb{N}}

\newcommand{\RR}{\mathbb R} 
 \newcommand{\G}{\mathcal G}
\newcommand{\T}{\mathcal T} 
\newcommand{\XX}{\mathcal I} \newcommand{\YY}{\mathcal Y}
\newcommand{\SX}{\mathcal S}
 
\newcommand{\x}{I}
\newcommand{\hh}{\mathcal{H}}
\newcommand{\F}{\mathcal{F}}
\newcommand{\Sg}{\mu}

\DeclareMathOperator{\R}{\mathbb{R}}

\providecommand{\nor}[1]{\left\lVert {#1} \right\rVert}
\providecommand{\scal}[2]{\left\langle{#1},{#2}\right\rangle}

\newcommand{\be}{\begin{equation}}
\newcommand{\ee}{\end{equation}}
\newcommand{\bt}{\begin{theorem}}
\newcommand{\et}{\end{theorem}}
\newcommand{\bd}{\begin{definition}}
\newcommand{\ed}{\end{definition}}
\newcommand{\bp}{\begin{proof}}
\newcommand{\ep}{\end{proof}}
\newcommand{\br}{\begin{remark}}
\newcommand{\er}{\end{remark}}
\newtheorem{ass}{Assumption}
\newtheorem{example}{Example}
\newtheorem{remark}{Remark}
\newtheorem{definition}{Definition}
\newtheorem{theorem}{Theorem}

\title{On Invariance and Selectivity in Representation Learning}

\author{Fabio Anselmi$^{1,2}$,Lorenzo Rosasco$^{1,2,3}$ and Tomaso Poggio$^{1,2}$}
\date{\today}

\begin{document}

\maketitle

\begin{abstract}
{We discuss data representation which can be learned automatically
from data, are invariant to
transformations, and at the same time selective, in the sense that
two points have the same representation only if they are one the
transformation of the other. The mathematical results here sharpen
some of the key claims of {\it i-theory} -- a recent theory of feedforward
processing in sensory cortex. \cite{anselmi2013unsupervised,AnselmiPoggio2014,MM2013}.}
\noindent 
The paper is submitted to Information and Inference Journal.
\end{abstract}

\footnotetext[1]{Center for Brains Minds and Machines, Massachusetts Institute of
Technology, Cambridge, MA 02139}
\footnotetext[2]{Laboratory for Computational Learning, Istituto Italiano di Tecnologia and Massachusetts Institute of Technology}
\footnotetext[3]{DIBRIS, Universit\'a degli studi di Genova, Italy, 16146}

\section{Introduction}

This paper considers the problem of learning {\em "good"} data
representation which can lower the need of labeled data (sample
complexity) in machine learning (ML).  Indeed, while current ML
systems have achieved impressive results in a variety of tasks, an
obvious bottleneck appears to be the huge amount of labeled data
needed.  This paper builds on the idea that data representation, which
are learned in an unsupervised manner, can be key to solve the
problem. Classical statistical learning theory focuses on supervised learning
and postulates that a suitable hypothesis space is given. In turn,
under very general conditions, the latter can be seen to be equivalent
to a data representation. In other words, data representation and how
to select and learn it, is classically not considered to be part of the learning problem, but
rather as a prior information. In practice ad hoc solutions are often
empirically found for each problem.

   The study in this paper is a step towards developing a theory of learning data representation.
   Our starting point is the intuition that, since many learning tasks
   are invariant to transformations of the data, learning invariant
   representation from ``unsupervised'' experiences can significantly
   lower the "size" of the problem, effectively decreasing the need of
   labeled data.  In the following, we formalize the above idea and
   discuss how such invariant representations can be learned.  Crucial
   to our reasoning is the requirement for invariant representations
   to satisfy a form of selectivity, broadly referred to as the
   property of distinguishing images which are not one the
   transformation of the other.  Indeed, it is this latter requirement
   that informs the design of non trivial invariant representations.
   Our work is motivated by a theory of cortex and in particular visual cortex \cite{MM2013}.

Data representation is a  classical concept in harmonic analysis and signal processing. 
Here  representations are typically designed on the basis of prior information assumed to be available.
More recently, there has been an effort to automatically learn
adaptive representation on the basis of data samples.  Examples in
this class of methods include so called dictionary learning
\cite{Tosic2011}, autoencoders \cite{Bengio2009} and metric learning
techniques (see e.g.  \cite{Xing03}). The idea of deriving invariant data
representation has been considered before. For example in the analysis
of shapes \cite{Kazhdan2003} and more generally in computational
topology \cite{Edelsbrunner10}, or in the design of positive definite
functions associated to reproducing kernel Hilbert spaces
\cite{burkhardt2007}.  However, in these lines of study the
selectivity properties of the representations have hardly been
considered.
The ideas in \cite{MallatGroupInvariantMain,soatto2009actionable} are
close in spirit to the study in this paper. In particular, the results
in \cite{MallatGroupInvariantMain} develop a different invariant and
stable representation within a signal processing framework. In
\cite{soatto2009actionable} an information theoretic perspective is
considered to formalize the problem of learning invariant/selective
representations.

In this work we develop a machine learning perspective closely
following computational neuroscience models of the information
processing in the visual cortex \cite{Hubel1965,Hubel1968,Riesenhuber1999}.  Our first and main
result shows that, for compact groups, representation defined by
nonlinear group averages can be shown to be invariant, as well as
selective, to the action of the  group. While invariance follows
from the properties of the Haar measure associated to the group,
selectivity is shown using probabilistic results that characterize a
probability measure in terms of one dimensional projections. This set
of ideas, which form the core of the paper, is then extended to
local transformations, and multilayer architectures.
These results  bear some understanding to the nature of certain deep architecture,
 in particular neural networks of the convolution type.

 The rest of the paper is organized as follows. We describe the
 concept of invariance and selective representation in Section
 \ref{InvSel} and their role for learning in Section \ref{LSC}. We
 discuss a family of invariant/selective representation for
 transformations which belong to compact groups in Section~\ref{sec:ildr}
 that we further develop in Sections~\ref{sec:pogildr} and ~\ref{sec:deepildr}.
Finally we conclude in Section~\ref{sec:end} with some final comments.
	
\section{Invariant and Selective Data Representations}\label{InvSel}

We next formalize and discuss the notion of {\em invariant and  selective} data representation,
which is the main focus of the rest of the paper.

 We model the data space as a (real separable) Hilbert space $\XX$ and denote by
 $\scal{\cdot}{\cdot}$ and  $\nor{\cdot}$  the  inner product and norm,  respectively.
 Example of data spaces are one dimensional signals (as in audio data), where we could let
 $\XX\subset L^{2}(\R)$, or two dimensional signals (such as images), where we could let
 $\XX\subset L^{2}(\R^2)$. After discretization, data can often be seen as vectors in
 high-dimensional Euclidean spaces, e.g.  $\XX=\R^d$.
The case of (digital) images  serves as a main example throughout the paper.

 A data representation is a map from the data space in  a suitable  representation space, that is
$$
\Sg:\XX\to {\mathcal F}.
$$
Indeed, the above concept appears under different names in various branch of pure and applied sciences, e.g. it is called
an encoding (information theory), a feature map (learning theory), a transform (harmonic analysis/signal processing)
or an embedding (computational geometry).

In this paper, we are interested in representations which are  invariant (see below)  to
suitable sets of  transformations. The latter  can be seen as a set of maps
$$\G\subset \{g~|~ g:\XX\to \XX\}.$$
Many interesting examples  of transformations have a group structure. Recall that a group
is a set endowed with a well defined {\em composition/multiplication} operation satisfying four  basic properties,
\begin{itemize}
\item closure:   $gg'\in \G$, for all $g,g'\in \G$
\item associativity:   $(gg')g''=g(g'g'')$, for all $g,g',g''\in \G$
\item identity:    there exists $\text{Id}\in \G$ such that $\text{Id}g=g\text{Id}=g$, for all $g\in \G$.
\item invertibility:   for all $g\in \G$  there exists $g^{-1}\in \G$ such that   $(gg^{-1})=\text{Id}$.
\end{itemize}
There are  different kind of groups. In particular, ``small" groups such as compact
(or locally compact, i.e.  a group  that admits a locally compact Hausdorff topology such that the group operations of composition and inversion are continuous.)  groups, or ``large" groups which are not locally compact.
In the case of images,  examples of locally compact groups include
affine transformations (e.g.   scaling, translations, rotations and their combinations) which can be thought of  as
suitable  {\em viewpoint} changes.  Examples of non locally compact groups are  diffeomorphisms,
which  can be thought  of as various kind of local or global  {\em deformations}.
\begin{example}
Let $\x\in L^2(\R)$. A basic example of group transformation is given by the translation group, which can be represented as a family of linear operators
$$
T_\tau: L^2(\R)\to L^2(\R), \quad T_\tau \x(p)=\x(p-\tau), \quad \forall p \in \R, \x\in \XX,
$$
for $\tau \in \R$.
Other basic examples of locally compact groups include scaling (the multiplication group) and affine transformations (affine group). Given a smooth map $d: \R\to \R$ a diffeomorphism can also be seen as a linear operator given by
$$
D_d: L^2(\R)\to L^2(\R), \quad D_d \x(p)=\x(d(p)), \quad  \forall p \in \R, \x\in \XX.
$$
Note that also in this case the representation is linear.
\end{example}
\noindent Clearly, not all transformations have a group structure-- think for example  of
images obtained from three dimensional rotations of an object. \\
Given the above premise, we next discuss, properties of data representation with respect to transformations.
We first add one remark about the notation.
\br[Notation: Group Action and Representation]
If $\G$ is a group and $\XX$ a set, the group action is  the map $(g,x)\mapsto g.x\in \XX$.
In the following, with an abuse of notation we will denote by $gx$ the group action. Indeed, when $\XX$
is a linear space, we also often denote by $g$ both a group element and its representation, so that $g$ can be identified with a linear operator. Throughout the article we assume the group representation to be unitary \cite{reed1978}.
\er

To introduce the notion of invariant representation, we recall that an orbit associated to an element $\x\in \XX$
is the  set $O_\x\subset \XX$ given by $O_\x=\{\x'\in \XX~|~ \x'=g\x, ~~ g \in \G \}$.  Orbits form a partition of $\XX$
in equivalence classes, with respect to the equivalence relation,
$$\x\sim \x'~~\Leftrightarrow ~~\exists ~g\in \G ~\text{such that} ~g\x=\x',$$
for all $\x,\x'\in \XX$.  We have the following definition.
\bd[Invariant Representation]\label{InvRep}
We say that a representation $\Sg$ is invariant with respect to $\G$ if
$$
\x\sim \x' \Rightarrow \Sg(\x)=\Sg(\x'),
$$
for all $\x, \x'\in \XX$.
\ed
In words, the above definition states that if two data points are one the transformation of the other, than they will have the same representation.
Indeed, if a representation $\Sg$ is invariant
$$\Sg(\x)=\Sg(g\x)$$
for all $\x\in \XX$, $g\in \G$.  Clearly, trivial invariant representations can be defined,
 e.g.  the constant function. This motivates a second requirement, namely selectivity.
\bd[Selective Representation]
We say that a representation $\Sg$ is selective with respect to $\G$ if
$$
\Sg(\x)=\Sg(\x') \Rightarrow \x\sim \x' ,
$$
for all $\x, \x'\in \XX$.
\ed
Together with invariance, selectivity asserts that two points have the same
representation {\em if and only} if they are one a transformation of the other.
Several comments are in order. First, the requirement of exact invariance as in Definition~\ref{InvRep}, seems desirable
for (locally) compact groups, but not for  non locally compact group such as diffeomorphisms.
In this case,   requiring a form of  stability to {\em small}  transformations seems to be natural, as it is more generally  to require stability to small perturbations, e.g. noise (see \cite{MallatGroupInvariantMain}).
Second, the concept of selectivity is natural  and  requires that no two orbits are mapped in the same representation.  It corresponds to an injectivity property of a representation  on the quotient space $\XX/\sim$. Assuming $\F$ to be endowed with a metric $d_\F$,  a stronger requirement  would be to characterize the metric embedding induced by $\Sg$, that is to control the ratio (or the deviation) of the distance of two representation and the distance of two orbits. Indeed, the problem of finding invariant and selective representation, is tightly related to the problem of finding an injective embedding of
 the quotient space $\XX/\sim$.

We next provide a discussion of the potential impact of invariant representations
on the  solution of subsequent learning tasks.

\section{From Invariance  to Low Sample Complexity}\label{LSC}

In this section we first  recall how the concepts of  data representation and hypothesis space are closely related, and how the sample complexity of a supervised problem can be characterized by the covering numbers  of the hypothesis space. Then, we discuss how invariant representations can lower the sample complexity of a supervised learning problem.

Supervised learning amounts to finding an input-output relationship on the basis of  a
{\em training} set of input-output pairs. Outputs can be scalar or vector valued,  as in regression, or  categorical, as in multi-category or multi-label classification, binary classification being a basic example.
The bulk of statistical  learning theory is devoted to study conditions under which learning problems can be  {\em solved}, approximately and up to a certain confidence,  provided a suitable hypothesis space is given.
A hypotheses space is a subset
$$\hh \subset \{ f~|~f:\XX\to \YY\},$$
of the set of all possible input output relations.
As we comment below, under very general assumptions {\em hypothesis spaces and  data representations are  equivalent concepts}.

\subsection{Data Representation and  Hypothesis Space}

Indeed, practically useful hypothesis spaces are typically endowed with a Hilbert space structure, since it is in this setting that most computational solutions can be developed.
A further natural requirement is for evaluation functions to be well defined and continuous.
This latter property allows to give a well defined meaning of the evaluation of a function at every points, a property which is arguably natural since we are interested in making predictions.
The  requirements of 1)
being   a Hilbert space of of functions and 2) have continuous evaluation functionals,
define so called reproducing kernel Hilbert spaces \cite{Ramm98onthe}.
Among other properties, these spaces of functions are characterized by the existence of
a feature map $\Sg:\XX\to \F$, which is a map from the data space into a feature space which is itself a Hilbert space.
Roughly speaking, functions in a RKHS $\hh$ with an associated feature map $\Sg$ can be seen as {\em hyperplanes} in the feature space,  in the sense that $\forall f\in \hh$, there exists $w\in \F$ such that
$$
f(\x)=\scal{w}{\Sg(\x)}_\F, \quad \forall \x\in \XX.
$$
The above discussion illustrates how, under mild assumptions, the choice of a hypothesis space is equivalent  to the choice of a data representation (a feature map).  In the next section, we recall how hypothesis spaces, hence data representation, are usually assumed to be given in statistical learning theory and  are characterized in terms of sample complexity.

\subsection{Sample Complexity in Supervised Learning}

Supervised statistical learning theory
characterizes the difficulty of a learning problem in terms of the "size" of the considered hypothesis space, as measured by suitable capacity measures. More precisely,  given a measurable loss function $V:\YY\times\YY\to [0,\infty)$,
for any measurable function $f:\XX\to \YY$ the expected error is defined as
$$
{\mathcal E}(f)=\int V(f(\x),y)d\rho(\x,y)
$$
where $\rho$ is a probability measure on $\XX\times \YY$.
Given a training set  $S_n=\{(\x_1,y_1),
\dots,\\ (\x_n,y_n) \}$ of input-output pairs sampled identically  and independently with respect to $\rho$, and a hypothesis space $\hh$, the goal of learning is to find an approximate solution $f_n=f_{S_n}\in \hh$ to the problem
$$
\inf_{f\in \hh} {\mathcal E}(f)
$$
%
%
The difficulty of a learning problem is captured by the following definition.
\bd[Learnability and Sample Complexity]
A hypothesis space $\hh$ is said to be learnable if, for all $\epsilon \in [0,\infty)$, $\delta \in [0,1]$, there exists $n(\epsilon, \delta, \hh)\in \N$
such that
\begin{equation}\label{sc}
\inf_{ f_n}\sup_{\rho}{\mathbb P}\left(  {\mathcal E}(f_n) - \inf_{f\in \hh}{\mathcal E}(f) \ge\epsilon \right)\le \delta .
\end{equation}
 The quantity $n(\epsilon, \delta, \hh)$ is called the sample complexity of the problem.
 \ed
 The above definition  characterizes the complexity of the learning problem associated to a hypothesis space $\hh$, in terms of the existence of
an algorithm that, provided with at least $n(\epsilon, \delta, \hh)$ training set points, can {\em approximately} solve the learning problem on $\hh$ with {\em accuracy} $\epsilon$
and {\em confidence} $\delta$.

The sample complexity associated to a hypothesis space $\hh$  can be derived   from
suitable notions of covering numbers, and related quantities, that characterize
the size of $\hh$.  Recall that, roughly speaking, the covering number $N_\epsilon$ associated to a (metric) space  is defined as the minimal number of $\epsilon$ balls needed to cover the space. The sample complexity can be shown \cite{Vapnik1982,CucSma02} to be proportional to the logarithm of the covering number, i.e.
$$
n(\epsilon, \delta, \hh)\propto  \frac{1}{\epsilon^2}\log \frac{N_\epsilon}{\delta}.
$$
As a basic example, consider  $\XX$ to be  $d$-dimensional and a hypothesis space of  linear functions
$$
f(\x)=\scal{w}{\x}, \quad \forall \x\in \XX, w\in \XX,
$$
so  that  the data representation is simply the identity.
Then the $\epsilon$-covering number of the set of linear functions with  $\nor{w}\le 1$ is given by
$$
N_\epsilon\sim \epsilon ^{-d}.
$$
If the input data lie in a subspace of dimension $s\le d$ then the
covering number of the space of linear functions becomes
$N_\epsilon\sim \epsilon ^{-s}$. In the next section, we further
comment on the above example and provide an argument to illustrate the
potential benefits of invariant representations.

\subsection{Sample Complexity of the Invariance Oracle}

Consider the  simple  example of a set of  images of $p\times p$ pixels each containing an   object   within a   (square) window of   $k\times k$ pixels and surrounded by a uniform background.
Imagine the object positions to be possibly anywhere in the image.
 Then it is easy to see that  as soon as objects are translated so that they  not overlap  we get an orthogonal subspace. Then, we see that there are $r^2=(p/k)^2$ possible subspaces of dimension $k^2$, that is the set of translated images can be seen as a distribution of vectors  supported within a ball in $d=p^2$ dimensions. Following the discussion in the previous section  the best  algorithm based on a linear hypothesis space will incur in a sample complexity proportional to  $d$. Assume now to have access to an {\em oracle} that can "register" each image so that  each object occupies the centered position. In this case, the  distribution of images is effectively  supported within a ball in $s=k^2$ dimensions and  the sample complexity is proportional to $s$ rather than $d$.  In other words a linear learning algorithm would need
$$
r^2=d/s
$$
less examples to achieve the  same accuracy. The idea is  that invariant representations
can act as an invariance oracle, and have the same impact on the sample complexity.
We add a few comments. First, while the above reasoning is developed for linear hypothesis space, a similar conclusion holds if non linear hypothesis spaces are considered. Second, one can see that the set of images obtained by translation is a low dimensional manifold,  embedded in a very high dimensional space. Other transformations, such as
small deformation, while being more complex, would have a much milder effect on the dimensionality of the embedded space. Finally, the natural question is how invariant representations  can be learned, a topic we address next.

\section{Compact Group Invariant  Representations}\label{sec:ildr}
Consider a set of transformations $\G$ which is a  locally compact group.
Recall that each locally  compact  groups has  a  finite measure naturally associated to it,
the so called Haar measure. The  key feature of the Haar measure  is its invariance to the group action, and in particular   for all measurable functions $f: \G\to \R$, and $g'\in \G$, it holds
$$
\int dg f(g)= \int dg f(g' g).
$$
The above equation  is reminding of  the invariance to translation of Lebesgue integrals and  indeed,  the Lebesgue measure can be shown to be the Haar measure associated to the translation group.  The invariance property of the Haar measure associated to a locally compact group, is  key to our development of invariant representation, as we describe next.


\subsection{Invariance via Group Averaging}

The starting point for deriving invariant representations is the following direct application  of the invariance property of the Haar measure.
\begin{theorem}
Let $\psi:\XX\to \R$ be a, possibly non linear, functional on $\XX$. Then,
the  functional defined by
\begin{equation}\label{GAve}
\Sg:\XX\to \R, \quad \quad \Sg(\x) =\int dg \psi(g\x), \quad \x\in \XX
\end{equation}
is invariant in the sense of Definition~\ref{InvRep}.
\end{theorem}
The functionals $\psi, \Sg$ can be thought to be measurements, or
features, of the data. In the following we are interested in
measurements of the form
\begin{equation}\label{NN}
\psi:\XX\to \R, \quad \quad \psi(\x) =\eta(\scal{g\x}{t}), \quad \x\in \XX, g\in \G
\end{equation}
where $t\in \T\subseteq \XX$ the set of unit vectors in $\XX$ and
$\eta:\R\to \R$ is a possibly non linear function.
As discussed in \cite{AnselmiPoggio2014},  the main motivation for considering measurements of the above form is their
interpretation in terms of biological or artificial neural networks,  see the following remarks.
\br[Hubel and Wiesel Simple and Complex Cells \cite{Hubel1962}]
  A measurement as in~\eqref{NN} can be interpreted as the output of a
  {\em neuron} which computes a possibly high-dimensional inner
  product with a template $t\in \T$. In this interpretation, $\eta$
  can be seen as a, so called, activation function, for which natural
  choices are sigmoidal functions, such as the hyperbolic tangent or
  rectifying functions such as the hinge.  The functional $\Sg$,
  obtained plugging~\eqref{NN} in~\eqref{GAve} can be seen as the
  output of a second neuron which aggregates the output of other
  neurons by a simple averaging operation. Neurons of the former kind
  are similar to simple cells, whereas neurons of the second kind are
  similar to complex cells in the visual cortex.
\er
\br[Convolutional Neural Networks \cite{lecun1995convolutional}]
  The computation of a measurement obtained plugging~\eqref{NN}
  in~\eqref{GAve} can also be seen as the output of a so called
  convolutional neural network where each neuron, $\psi$ is performing
  the inner product operation between the input, $I$, and its synaptic
  weights, $t$, followed by a pointwise nonlinearity $\eta$ and a
  pooling layer.
\er

 A second, reason to consider measurements of the form~\eqref{NN} is  computational and, as shown later,   have direct implications for learning. Indeed,  to compute an invariant feature, according to~\eqref{GAve} it is necessary to  be able to compute the action of any  element $\x\in \XX$ for which
we wish to compute the invariant measurement. However, a simple observation suggests an alternative strategy. Indeed, since  the group representation is  unitary, then
$$
\scal{g\x}{\x'}=\scal{\x}{g^{-1}\x'}, \quad \forall \x,\x'\in \XX
$$
so that in particular we can compute $\psi$ by considering
\be\label{GAveFeat} \psi(\x) =\int dg \eta(\scal{\x}{gt}), \quad \forall \x\in
\XX, \ee where we used the invariance of the Haar measure. The above
reasoning implies that an invariant feature can be computed for any
point provided that for $t\in \mathcal T$, the sequence $gt$, $g\in
\G$ is available.  This observation has the following interpretation:
if we view a sequence $gt$, $g\in \G$, as a "movie" of an object
undergoing a family of transformations, then the idea is that
invariant features can be computed for any new image provided that a
movie of the template is available.

While group averaging provides a natural way to tackle the problem of invariant representation,
it is not clear how a family of invariant measurements can be ensured to be  selective.
Indeed, in the case of compact groups selectivity
can be provably characterized using a probabilistic argument summarized in the following three steps:
\begin{enumerate}
\item A unique probability distribution can be naturally associated to each orbit.
\item Each such  probability distributions can be characterized in terms of  one-dimensional projections.
\item One dimensional probability distributions are easy to characterize, e.g.  in terms of their
cumulative distribution or their moments.
\end{enumerate}
We note in passing that the above development, which we describe in
detail next, naturally provides as a byproduct indications on how the
non linearity in~\eqref{NN} needs to be chosen and thus gives insights on
the nature of the {\em pooling} operation.

\subsection{A Probabilistic Approach to Selectivity}

Let $\XX=\RR^d$,  and  ${\mathcal P}(\XX)$
the space of probability measures on $\XX$. Recall that for any compact group, the Haar measure is finite, so that, if appropriately normalized, it correspond to a probability measure.
\begin{ass}
In the following we assume $\G$ to be Abelian and compact and the corresponding Haar measure to be normalized.
\end{ass}
The first step in our reasoning is the following definition.
\bd[Representation via Orbit Probability]\label{ProbRep}
For all $\x\in \XX$, define the random variable
$$
Z_\x:(\G,dg)\to \XX, \quad Z_\x(g)=g\x, \quad \forall g\in \G,
$$
with law
 $$
 \rho_\x(A)=
 \int_{Z_\x^{-1}(A)} dg,
 $$
 for all measurable sets $A\subset \XX$.
Let
 $$
 P:\XX\to {\mathcal P}(\XX), \quad P(\x)=\rho_\x, \quad \forall \x\in \XX.
 $$
 \ed
The map $P$ associates to each point a corresponding  probability distribution.
From the above definition we see that we are essentially viewing an orbit as a distribution of points, and mapping each point in one such distribution. Then we have the following result.
 \bt\label{PRep}
 For all $\x,\x'\in \XX$
 \be
\x\sim \x'~~ \Leftrightarrow ~~ P(\x)=P(\x').
 \ee
\et
\bp
We first prove that $\x\sim \x' \;\Rightarrow\; \rho_{\x}=\rho_{\x'}$.
Recalling that if $\mathcal{C}_{c}(\mathcal{\x})$ is the set of continuous functions on $\mathcal{I}$ with compact support,  $\rho_{I}$ can be alternatively defined as the unique probability distribution such that
\begin{equation}\label{alternativelaw}
\int\; f(z)d\rho_{I}(z) = \int\; f(Z_{I}(g))dg, \quad  \forall f\in \mathcal{C}_{c}(\mathcal{\x}).
\end{equation}
Therefore $\rho_{\x}=\rho_{\x'}$ if and only if for any $f\in \mathcal{C}_{c}(\mathcal{\x})$, we have $\int_{\mathcal{G}}f(Z_{\x}(g))dg = \int_{\mathcal{G}}f(Z_{\x'}(g))dg$ which follows immediately by a change of variable and invariance of the Haar measure:
$$
\int_{\mathcal{G}}f(Z_{\x}(g))dg = \int_{\mathcal{G}}f(g\x)dg = \int_{\mathcal{G}}f(g\x')dg =
\int_{\mathcal{G}}f(g\tilde g\x)dg =  \int_{\mathcal{G}}f(\hat g\x)d\hat g
$$
To prove that $\rho_{\x}=\rho_{\x'}\;\Rightarrow\; \x\sim \x'$, note that $\rho_{\x}(A)-\rho_{\x'}(A)=0$ for all measurable sets $A\subseteq \XX$ implies in particular that the support of the probability distributions of $\x$ has non null intersection on a set of non zero measure. Since the support  of the distributions $\rho_\x,\rho_{\x'}$ are exactly the orbits associated to $\x, \x'$ respectively, then the orbits coincide, that is $\x\sim \x'$.
\ep

The above result shows that an invariant representation can be defined considering the probability distribution naturally associated to each orbit, however  its computational realization would require dealing with high-dimensional distributions.  Indeed, we next show that the above representation can be further developed to consider  only   probability distributions on the real line.

\subsubsection{Tomographic Probabilistic Representations}

We  need to introduce some notation and definitions.  Let  $\T=\SX$,  the unit sphere in $\XX$, and
let ${\mathcal P}(\RR)$ denote the set of  probability measures on the real line.
For each $t\in \T$, let
$$
\pi_t:\XX\to \RR, \quad \pi_t(\x)=\scal{\x}{t},\quad \forall \x\in \XX.
$$
 If $\rho \in  {\mathcal P}(\XX)$, for all $t\in \T$ we denote by   $\rho^t\in  {\mathcal P}(\RR)$
 the random variable with law   given by
$$
\rho^t(B)=
\int_{\pi_t^{-1}(B)} d\rho,
$$
 for all measurable sets $B\subset \RR$.
 \bd[Radon Embedding]\label{RadRep}
Let ${\mathcal P}(\RR)^{\T}=\{h~|~h:\T\to{\mathcal P}(\RR) \}$ and  define
$$
R: {\mathcal P}(\XX) \to {\mathcal P}(\RR)^{\T}, \quad R(\rho)(t)= \rho^t ,\quad \forall \x\in \XX.
$$
\ed
\noindent The above map associates to each probability distribution a (continuous) {\em family} of probability distributions
on the real line defined by one dimensional projections ({\em tomographies}). Interestingly, $R$ can   be shown to be a generalization  of the Radon Transform to probability distributions \cite{Boman2009}. We are going to use it to define the following data representation.
\bd[TP Representation]\label{TPRep}
We define the Tomographic Probabilistic (TP) representation as
$$
\Psi:\XX\to {\mathcal P}(\RR)^{\T}, \quad \Psi=R\circ P,
$$
with $P$ and $R$ as in Definitions ~\ref{ProbRep},~\ref{RadRep}, respectively.
\ed
\noindent The TP representation is obtained  by first mapping each point  in the distribution supported on  its orbit and then in a  (continuous) family of corresponding one dimensional distributions.
The following result characterizes the invariance/selectivity property of the TP representation.
\bt\label{RadRepThm}
Let $\Psi$ be the TP representation in Definition~\ref{TPRep},  then for all $\x,\x'\in \XX$
 \be
\x\sim \x'~~ \Leftrightarrow ~~ \Psi(\x)=\Psi(\x').
 \ee
\et
\noindent
The proof of the above result
is obtained combining Theorem~\ref{PRep} with the following well known result, characterizing probability distributions in terms of their one dimensional projections.
\bt[Cramer-Wold \cite{CramerWold1936}]\label{CW}
For any $\rho, \gamma\in {\mathcal P}(\XX)$, it holds
\be
\rho=\gamma~~ \Leftrightarrow ~~ \rho^t=\gamma^t, \quad\forall t ~\in \SX.
\ee
\et
\noindent
Through the  TP representation,  the problem of finding invariant/selective representations
reduces to the study of one dimensional distributions, as we discuss next.

\subsubsection{CDF Representation}\label{Sec:CDFRep}

A natural way to describe a one-dimensional  probability distribution is to consider the associated cumulative distribution function (CDF). Recall that if $\xi:(\Omega, p)\to \R$ is a random variable with law $q\in {\mathcal P}(\RR)$, then the associated CDF is given by
\be\label{CDF}
f_q(b)= q((\infty, b])=\int d p(a)
H( b-\xi(a))
, \quad b\in \R,
\ee
where where $H$ is the Heaviside step function.
Also recall that the CDF uniquely defines a probability distribution since, by the Fundamental Theorem of Calculus, we have
$$
\frac{d}{db}f_q(b)=\frac{d}{db}\int d p(a)
H( b-\xi(a))
=\frac{d}{db}\int_{-\infty}^{b} d p(a) = p(b).
$$
We consider the following map.
\bd[CDF Vector Map]\label{CDFRep}
Let ${\mathcal F}(\RR)=\{h~|~h:\RR\to [0, 1]\}$, and
$${\mathcal F}(\RR)^\T=\{ h~|~h:\T\to {\mathcal F}(\RR)\}.$$
Define
$$
F: {\mathcal P}(\RR)^\T  \to {\mathcal F}
(\RR)^\T, \quad F(\overline{\gamma})(t)= f_{\overline{\gamma}^t}
$$
for $\overline \gamma \in {\mathcal P}(\RR)^{\mathcal{T}}$ and where we let $\overline{\gamma}^t=\overline{\gamma}(t)$ for
all $t\in \T$.
\ed
\noindent
The above map associates to a family of probability distributions on the real line their corresponding CDFs. We can then define the following representation.
\bd[CDF Representation]
Let
$$
\Sg:\XX\to {\mathcal F}(\RR)^{\T}, \quad \Sg=F\circ R\circ P,
$$
with $F$,$P$ and $R$ as in Definitions~\ref{CDFRep},~\ref{ProbRep},~\ref{RadRep}, respectively.
\ed
\noindent
Then, the following result holds.
\bt\label{CDFrepT}
For all $\x\in \XX$ and $t\in \T$
\be\label{CDFReRep}
\Sg^t(\x)(b)=\int dg \eta_b(\scal{\x}{gt}), \quad b\in \R,
\ee
where we let $\Sg^t(\x)=\Sg(\x)(t)$ and,  for all $b\in \R$, $\eta_b:\R\to \R$, is given by
$\eta_b(a)= H(b-a), ~~a\in \R$. Moreover, for all $\x,\x'\in \XX$
$$
\x\sim \x'~~ \Leftrightarrow ~~ \Sg(\x)=\Sg(\x').
$$
\et
\bp
The proof follows noting that $\Sg $ is the composition of the one to one maps $F,R$ and a map $P$ that is one to one w.r.t. the equivalence classes induced by the group of transformations $G$. Therefore $\Sg$ is one to one w.r.t. the equivalence classes i.e. $\x\sim \x'~~ \Leftrightarrow ~~ \Sg(\x)=\Sg(\x')$.
\ep

We note that, from a direct comparison,  one can see that~\eqref{CDFReRep} is of the form~\eqref{GAveFeat}. Different measurements correspond to different choices of the threshold $b$.

\br\label{Rem:MomRep}[Pooling Functions: from CDF to Moments and Beyond]
The above reasoning suggests that a principled choice for the non linearity in~\eqref{GAveFeat}
is a step function, which in practice could be replaced by a smooth approximation such a sigmoidal function.
Interestingly, other choices of non linearities could be considered. For example, considering different powers would
yield information on the  moments of the distributions (more general non linear function than powers
would yield generalized moments). This latter point of view is discussed in some detail in Appendix~\ref{Sec:Mom}.
\er

\subsection{Templates Sampling and Metric Embedings}

We next discuss what happens if only a finite number of (possibly random) templates are available.
In this case, while invariance can be ensured, in general we cannot expect  selectivity to be preserved.
However,  it is possible to show that the representation is {\em  almost} selective (see below)  if a sufficiently large number  number of templates is available.

Towards this end we introduce a metric structure on the representation space.
Recall that if $\rho, \rho'\in {\mathcal P}(\R)$ are two probability distributions on the real line and $f_\rho, f_{\rho'}$ their cumulative distributions functions, then the uniform Kolmogorov-Smirnov (KS)  metric is
induced by the uniform norm of the  cumulative distributions that is
$$
d_\infty(f_\rho,f_{\rho'})=\sup_{s\in \R} |f_\rho(s)-f_{\rho'}(s)|,
$$
and takes values in $[0,1]$. Then, if $\Sg$ is the representation  in~\eqref{CDFReRep}
we can consider the metric
\be\label{SlicedMetric}
d(\x,\x')= \int d u(t) d_\infty (\Sg^t(\x),\Sg^t(\x'))
\ee
where $u$ is the (normalized) uniform measure on the sphere $\SX$.  We note that, theorems~\ref{CW} and~\ref{CDFrepT} ensure that ~\eqref{SlicedMetric} is a well defined metric on the quotient space induced by the group transformations,  in particular
$$
d(\x,\x')=0\Leftrightarrow \x\sim\x'.
$$
If we consider the case in which only a finite set $\T_k= \{t_1,\dots, t_k\}\subset \SX$ of $k$ templates is available,
each point is mapped in a finite sequence of probability distributions or CDFs and~\eqref{SlicedMetric} is replaced by
\be\label{EmpSlicedMetric}
\widehat d(\x,\x')= \frac 1 k \sum_{i=1}^k d_\infty (\Sg^{t_i}(\x),\Sg^{t_i}(\x'))
\ee
Clearly, in this case we cannot expect to be able to discriminate every pair of points, however we have the following result.
\bt
\label{akaJL}
Consider $n$ images $\XX_n$ in $\XX$. Let $k \ge
\frac{2}{c\epsilon^2} \log {\frac{n}{\delta}}$, where $c$ is a constant.  Then
with probability $1 - \delta^2$,
\begin{equation} \label{CWJL}
|d(\x,\x') - \widehat{d}(\x,\x')|\le\epsilon.
\end{equation}
for all  $\x,\x'\in \XX_n$.
\et
\bp
The proof follows from a direct application of H\"oeffding's inequality and a union bound.
Fix $\x,\x'\in \XX_{n}$. Define the real random
variable $Z:\SX \to [0,1]$,
$$
Z(t_i)=
d_\infty (\Sg^{t_i}(\x),\Sg^{t_i}(\x')),\quad i=1, \dots, k.
$$
From the definitions it follows that
$\nor{Z}\le 1$  and $\mathbb E (Z)=d(\x,\x')$. Then, H\"oeffding
inequality implies
$$ |d(\x, \x') -\widehat{d}(\x,\x')|= |\frac 1 k \sum_{i=1}^{k} \mathbb E (Z) - Z(t_{i}) | \ge \epsilon,$$
with probability at most $2e^{-\epsilon^2k}$.  A union bound implies
that the result holds uniformly on $\XX_{n}$ with probability at least
$n^22e^{-\epsilon^2k}$. The proof is concluded setting this
probability to $\delta^2$ and taking $k \ge \frac{2}{c\epsilon^2} \log
{\frac{n}{\delta}}$.
\ep

We note that, while we considered the KS
metric for convenience, other metrics over probability distributions
can be considered.  Also, we note that a natural further question is how discretization/sampling
of the group affects the representation. The above reasoning could be extended to yield results in this latter case.
Finally, we note that, when compared to classical
results on distance preserving embedding, such as Johnson
Linderstrauss Lemma \cite{JL84}, Theorem~\ref{EmpSlicedMetric} only
ensures distance preservation up to a given accuracy which increases
with a  larger number of projections.  This is hardly
surprising, since the problem of finding suitable embedding for
probability spaces is known to be considerably harder than the
analogue problem for vector spaces \cite{AndoniBIW09}. The question of
how devise strategies to define distance preserving embedding is an
interesting open problem.

\section{Locally Invariant and Covariant Representations}\label{sec:pogildr}

We consider the case where a representation is given by collection of {\em "local"} group averages, and refer to this situation as the partially observable group (POG) case. Roughly speaking, the idea is that this kind of measurements can be invariant to sufficiently small transformations, i.e. be locally invariant. Moreover, representations given by collections of POG averages can be shown to be
{\em covariant} (see section \ref{Pogrep} for a definition).

\subsection{Partially Observable Group Averages}

\begin{figure}
\centering
\includegraphics[width= 10cm]{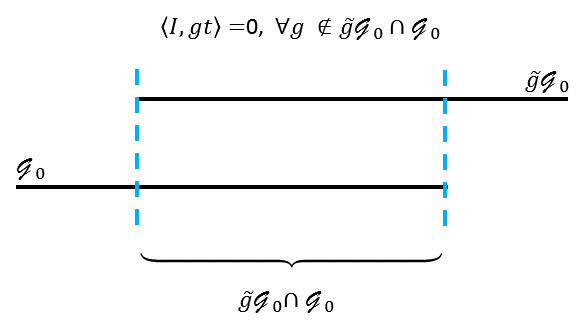}
\caption{A sufficient condition for invariance for locally compact groups: if $\scal{gI}{t}=0$ for all $g \in \tilde  g \G_0 \Delta \G_0$, the integral of $\eta_{b}\scal{I}{gt}$ over $\G_0$ or $\tilde g \G_{0}$ will be equal.\label{LCG_condition}}
\end{figure}

For a subset  $\G_0\subset \G$  consider a POG measurement of the form
\be\label{pogmeas}
\psi(\x)= \int_{\G_0} dg  \eta(\scal{\x}{gt}).
\ee
The above quantity can be interpreted as the "response" of a cell that can perceive visual stimuli within a "window" (receptive field)
of size $\G_0$.  A POG measurement corresponds to a local group average restricted to a subset of transformations $\G_0$.
Clearly,  such a measurement will not in general be invariant. Consider a POG measurement on
a transformed point
$$
\int_{\G_0} dg  \eta(\scal{\tilde g \x}{gt})=\int_{\G_0} dg  \eta(\scal{\x}{\tilde g^{-1}gt})=\int_{\tilde g \G_0} dg  \eta(\scal{\x}{gt}).
$$
If we compare the POG measurements on the same point with and without
a transformation, we have \be\label{diff} |\int_{\G_0} dg
\eta(\scal{\x}{gt})- \int_{\tilde g \G_0} dg \eta(\scal{\x}{gt})|.
\ee While there are several situations in which the above difference
can be zero, the intuition from the vision interpretation is that the
same response should be obtained  if a sufficiently small object does
not move (transform) too much with respect to the receptive field
size.  This latter situation can be described by the assumption that
the function
$$
h:\G\to \R, \quad h(g)=\eta(\scal{\x}{gt})
$$
is zero outside of the intersection of $\tilde g \G_0\cap
\G_0$. Indeed, for all $\tilde g \in \G$ satisfying this latter
assumption, the difference in~\eqref{diff} would clearly be zero.  The
above reasoning results in the following theorem.  \bt Given $\x\in
\XX$ and $t\in \T$, assume that there exists a set $\tilde \G\subset
\G$ such that, for all $\tilde g \in \tilde \G$, \be\label{loc}
\eta(\scal{\x}{gt})=0 \quad \forall g \notin \tilde g \G_0 \cap \G_0.
\ee Then for $\tilde g \in \tilde \G$
$$
\psi(I)=\psi(\tilde g \x),
$$
with $\psi$ as in~\eqref{pogmeas}.
\et
We add a few comments. First, we note that condition~\eqref{loc} can be weakened requiring only
$\eta(\scal{\x}{gt})=0$ for all $g \in \tilde g \G_0 \Delta \G_0$, where we denote by $\Delta$ the
symmetric difference  of two sets ($ A\Delta B = (A\cup B)/ (A\cap B)$ with $A,B$ sets). Second, we note that if the non linearity
$\eta$ is zero only in zero, then we can rewrite condition~\eqref{loc} as
$$
\scal{\x}{gt}=0, \quad  \forall g \in   \tilde g \G_0 \Delta \G_0.
$$
Finally, we note that the latter expression has a simple
interpretation in the case of the translation group. In fact, we can
interpret~\eqref{loc} as a spatial localization condition on the image $\x$
and the template $t$ (assumed to be positive valued functions), see
Figure~\ref{LCG_condition}.  We conclude with the following remark.
\br[Localization Condition and V1] Regarding the localization condition
discussed above, as we comment elsewhere
\cite{anselmi2013unsupervised}, the fact that a template needs to be
localized could have implications from a biological modeling
standpoint. More precisely, it could provides a theoretical foundation
of the Gabor like shape of the responses observed in V1 cells in the
visual cortex  \cite{Poggio2013,anselmi2013unsupervised,MM2013}.
\er
\br[More on the Localization Condition]
From a more mathematical
point of view, an interesting question is  about conditions under
which  whether the
localization condition~\eqref{loc} is also necessary
rather than only sufficient.
\er

%
%
%

\subsection{POG Representation}\label{Pogrep}

For all $ \bar g \in G$, let   $\bar g\G_0=\{g\in \G~|~ g=\bar g g',\quad g'\in \G_0\}$, the collection of "local" subsets of the group obtained from the subset $\G_0$.
Moreover, let
$$
V=\int_{\G_0}dg.
$$
Clearly, by the invariance of the measure, we have $\int_{\bar g  \G_0}dg=V$, for all $\bar g \in \G$.
Then,  for all $\x\in \XX$, $ \bar g \in \G$, define the random variables
\be\label{Zxg}
Z_{\x,\bar{g}}:\bar g \G_0\to \XX, \quad Z_{\x,\bar{g}}(g)=g\x, \quad g \in  \bar g \G_{0},
\ee
with laws
 $$
 \rho_{\x, \bar g}(A)=
\frac 1 V  \int_{Z_{\x, \bar g}^{-1}(A)}
dg,t
 $$
for all measurable sets $A\subset \XX$.
For each $\x\in \XX$, $ \bar g \in \G$,
the measure $\rho_{\x, \bar g}$ corresponds to the distribution on the fraction of the orbit corresponding to the observable group subset $\bar g\G_0$. Then we can represent each point with a collection of POG distributions.

\bd[Representation via POG Probabilities]\label{POGProb}
Let ${\mathcal P}(\XX)^{\G}=\{h~|~h:\G\to{\mathcal P}(\XX) \}$ and  define
$$
 \bar P:\XX\to {\mathcal P}(\XX)^\G, \quad \bar  P(\x)(g)=\rho_{\x, g} \quad \forall \x\in \XX,g\in \G
 $$
\ed
\noindent
Each point is mapped in the  collection of distributions obtained considering all possible fractions of the orbit corresponding to $\bar g \G_0$, $\bar g \in \G$.  Note that, the action of an element $\tilde g \in \G$ of the group on the POG probability representation is given by
$$
\tilde g \bar P(\x)(g)=  \bar P(\x)(\tilde g g)
$$
for all $g\in \G$. The following result holds.
\bt\label{P-POGRep}
   Let $\bar P$ as in Definition~\eqref{POGProb}. Then
 for all $\x,\x'\in \XX$ if
 \be
\x\sim \x'~~ \Rightarrow ~~ \exists \tilde g\in \G~\text{such that}~\bar P(\x')=  \tilde g \bar P(\x).
 \ee
Equivalently, for all $\x,\x'\in \XX$ if
$$
\x'=\tilde g \x
 $$
then
\begin{equation}\label{Covariance}
\bar P(\x')(g)= \bar P(\x)(g\tilde g), \quad \forall g\in \G.
\end{equation}
i.e. $\bar P$ is \emph{covariant}.
\et
\bp
The proof follows noting that $\rho_{\x',\bar g} = \rho_{\x,\bar g \tilde g}$ holds  since, using the same characterization of $\rho$ as in \eqref{alternativelaw},we have that for any $f\in \mathcal{C}_{c}(\mathcal{I})$
$$
\int_{\bar g \G_{0}}f(Z_{\x',\bar g}(g))dg = \int_{\bar g \G_{0}}f(g\x')dg = \int_{\bar g \G_{0}}f(g\tilde g \x)dg = \int_{\bar g \G_{0}\tilde g}f(g\x)dg
$$
where we used  the invariance of the measure.
\ep

Following the reasoning in the previous sections and recalling Definition~\ref{RadRep}, we consider the mapping given by one dimensional projections  (tomographies) and corresponding representations.
\bd[TP-POG Representation]\label{POGRad}
Let ${\mathcal P}(\R)^{\G\times \T}=\{h~|~h:\G\times \T\to{\mathcal P}(\R) \}$ and  define
$$
 \bar R : {\mathcal P}(\XX) ^\G \to {\mathcal P}(\R)^{\G\times \T},
 \quad \bar  R(h)(g,t)=R(h(g))(t)=h^t(g),
  $$
for all $h\in  {\mathcal P}(\XX) ^\G,g\in \G, t\in \T$. Moreover,  we  define the Tomographic Probabilistic POG representation as
$$
\bar \Psi:\XX\to {\mathcal P}(\RR)^{\G\times\T}, \quad \bar \Psi=\bar R\circ \bar P,
$$
with $\bar P$ as in Definition~\ref{POGProb}.
\ed
We have the following result:
\bt\label{Psicov}
The representation $\bar \Psi $ defined in \ref{POGRad} is covariant, i.e. $\bar \Psi (\tilde g \x)(g) = \bar \Psi (\x)(\tilde g g) $.
\et
\bp
The map  $\bar \Psi = \bar R\circ \bar P $ is covariant if both $\bar R$ and $\bar P$ are covariant. The map $\bar P$ was proven to be covariant in Theorem \ref{P-POGRep}. We then need to prove the covariance of $\bar R$ i.e. $\tilde g \bar R(h)(g,t)=\bar R(h)(\tilde g g,t)$ for all $h\in {\mathcal P}(\XX) ^\G $. This follows from
\begin{eqnarray*}
\bar R(\tilde g h)(g,t) = R( \tilde g h(g))(t) = R( h(\tilde g g))(t) = R(h)(\tilde g g,t).
\end{eqnarray*}
\ep

The TP-POG representation is obtained  by first mapping each point  $\x$ in the family of distributions $\rho_{\x,g}$, $g\in \G$ supported on the  orbit fragments corresponding to POG and then in a  (continuous) family of corresponding one dimensional distributions
$\rho^t_{\x,g}$, $g\in \G$, $t\in \T$. Finally, we can consider  the representation obtained representing each distribution via the corresponding CDF.
\bd[CDF-POG Representation]\label{CDFPOG}
Let ${\mathcal F}(\R)^{\G\times \T}=\{h~|~h:\G\times \T\to{\mathcal F}(\R) \}$ and  define
$$
 \bar F : {\mathcal P}(\XX)^{\G\times \T} \to {\mathcal P}(\R)^{\G\times \T},
 \quad \bar  F(h)(g,t)=F(h(g,t))=f_{h(g,t)},
  $$
for all $h\in  {\mathcal P}(\XX) ^{\G\times \T}$ and $g\in \G, t\in \T$. Moreover,   define the CDF-POG representation as
$$
\bar \Sg:\XX\to {\mathcal F}(\RR)^{\G\times\T}, \quad \bar \Sg=\bar F\circ\bar R\circ \bar P,
$$
with $\bar P$,$\bar F$ as in Definition~\ref{POGProb}, ~\ref{POGRad}, respectively.
\ed
It is easy to show that
\be\label{CDFPOGEx}
\Sg_{\bar g,t}(\x)(b)=\int_{\bar g \G_{0}} \eta_b(\scal{\x}{gt})dg.
\ee
where we let $\Sg_{\bar g,t}(\x)=\Sg(\x)(\bar g,t)$.

\section{Further Developments:  Hierarchical Representation}\label{sec:deepildr}

In this section we discuss some further developments of the framework presented in the previous section.
In particular, we sketch how multi-layer (deep) representations can be obtained abstracting and iterating
the basic ideas introduced before.


Hierarchical representations, based on multiple layers of
computations, have naturally arisen from models of information
processing in the brain \cite{Fukushima1980,Riesenhuber1999}. They
have also been critically important in recent machine learning
successes in a variety of engineering applications, see
e.g. \cite{sermanet-iclr-14}.  In this section we address the question
of how  to generalize the framework previously introduced to
consider multi-layer representations.

Recall that the basic idea for building invariant/selective
representation is to consider local (or global) measurements of the
form
\be\label{loc2} \int_{\G_0}\eta(\scal{\x}{gt})dg, \ee
with
$\G_0\subseteq \G$.  A main difficulty to iterate this idea is that,
following the development in previous sections, the
representation~\eqref{CDFPOG}-\eqref{CDFPOGEx}, induced by collection
of (local) group averages, maps the data space $\XX$ in the space
${\mathcal P}(\R)^{\G\times\T}$. The latter space lacks an inner
product as well as natural linear structure needed to define the
measurements in~\eqref{loc2}.  One possibility to overcome this
problem is to consider an embedding in a suitable Hilbert space. The
first step in this direction is to consider an embedding of the
probability space ${\mathcal P}(\R)$ in a (real separable) Hilbert
space $\hh$. Interestingly, this can be achieved considering a
variety of reproducing kernels over probability distributions, as we
describe in Appendix~\ref{RKHSProb}.  Here we note that if
$\Phi:{\mathcal P}(\R)\to \hh$ is one such embeddings, then we could
consider a corresponding embedding of ${\mathcal P}(\R)^{\G\times\T}$
in the space
$$
L^2(\G\times\T, \hh)=\{ h: \G\times\T\to \hh~|~ \int \nor{h(g,t)}^2dg du(t)\}
$$
where $\nor{\cdot}_\hh$ is the norm induced by the  inner product $\scal{\cdot}{\cdot}_\hh$ in $\hh$ and $u$ is the uniform measure on the sphere $\SX\subset \XX$. The space $L^2(\G\times\T, \hh)$ is endowed with the inner product
$$
\scal{h}{h'}_{\mathcal{H}}=\int \scal{h(g,t)}{h'(g,t)}_{\hh}^2dg du(t),
$$
for all $h,h'\in L^2(\G\times\T, \hh)$, so that the corresponding norm is exactly
$$
\nor{h}^{2}_{\mathcal{H}}=\int \nor{h(g,t)}^2dg du(t).
$$
The embedding of ${\mathcal P}(\R)^{\G\times\T}$ in  $L^2(\G\times\T, \hh)$ is simply given by
$$
J_\Phi:{\mathcal P}(\R)^{\G\times\T}\to L^2(\G\times\T, \hh), \quad J_\Phi(\rho)(g,t)= \Phi(\rho(g,t)) \quad \text{i.e.}
$$
for all $\rho\in {\mathcal P}(\R)^{\G\times\T}$.
Provided with above notation we have the following result.
\bt
The representation defined by
\be\label{QRep}
\bar Q:\XX\to L^2(\G\times\T, \hh), \quad \bar Q=J_\Phi \circ \bar \Psi.
\ee
with $\bar\Psi$ as in Definition~\ref{POGRad}, is covariant, in the sense that,
$$
\bar Q(g\x)=g\bar Q(\x)
$$
for all $\x\in \XX$, $g\in \G$.
\et
\bp
The proof follows checking that by definition both $\bar R $ and $J_\Phi$ are covariant and using Theorem~\ref{P-POGRep}. The fact that $\bar R$ is covariant was proven in Th. \ref{Psicov}. The covariance of $J_\Phi$, i.e. $\tilde g J_{\Phi}(h)(g,t) =  J_{\Phi}(h)(\tilde g g,t)$ for all $h\in {\mathcal P}(\R)^{\G\times\T}$,  follows from
\begin{eqnarray*}
&&J_{\Phi}(\tilde g h)(g,t) = \Phi(\tilde g h(g,t)) = \Phi(h(\tilde g g,t)) =
 J_{\Phi}( h)(\tilde g g,t).
\end{eqnarray*}
Now since $\bar P$ was already proven covariant in Th. \ref{P-POGRep} we have that, being $\bar Q = J_{\Phi}\circ \bar R \circ \bar P$ composition of covariant representations, $\bar Q$ is covariant i.e. $\tilde g \bar Q(\x) =  \bar Q(\tilde g \x)$.
\ep

Using the above definitions a {\em second layer} invariant measurement can be defined considering,
\be\label{deeploc}
v:\XX\to \R, \quad
v(\x)=\int_{\G_0}\eta(\scal{\bar Q(x)}{ g \tau }_2)dg
\ee
where $\tau \in L^2(\G\times\T, \hh)$ has unit norm.

We add several comments.  First, following the analysis in the
previous sections Equation~\eqref{deeploc} can be used to define
invariant (or locally invariant) measurements and hence
representations defined by collections of measurements. Second, the
construction can be further iterated to consider multi-layer
representations, where at each layer an intermediate representation is
obtained considering "distributions of distributions".  Third,
considering multiple layers naturally begs the question of how the
number and properties of each layer affect the properties of the
representation. Preliminary answers to these questions are described
in
\cite{anselmi2013unsupervised,AnselmiPoggio2014,leibo2014modularity,Poggio2013}. A
full mathematical treatment is beyond the scope of the current paper
which however provides a formal framework to tackle them in future work.

\section{Discussion}\label{sec:end}
Motivated by the goal of characterizing good data representation that
can be learned, this paper studies the mathematics of an approach
to learn data representation that are invariant and selective to
suitable transformations. While invariance can be proved rather
directly from the invariance of the Haar measure associated with the
group, characterizing selectivity requires a novel probabilistic
argument developed in the previous sections.\\
\noindent
Several extensions of the theory are natural and have been sketched with  preliminary results in
\cite{anselmi2013unsupervised,AnselmiPoggio2014,leibo2014modularity,Poggio2013}.
The main directions that need a rigorous theory extending the results
of this paper are:
\begin{itemize}
\item Hierarchical architectures. We described how the theory can be
  used to analyze local invariance properties, in particular for
locally compact groups. We described covariance properties. Covariant
layers can integrate representations that are locally invariant into
representations that are more globally invariant.
\item Approximate invariance for transformations that are not
  groups. The same basic algorithm analyzed in this paper is used to
  yield approximate invariance, provided the templates transforms as
  the image, which requires the templates to be tuned to specific object classes.
\end{itemize}
We conclude with a few general remarks connecting our paper with this special issue
on deep learning and especially with an eventual theory of such networks.\\
\noindent
{\it Hierarchical architectures of simple and complex units}. Feedforward architecture with
$n$ layers, consisting of dot products and nonlinear pooling functions, are
quite general computing devices, basically equivalent to Turing
machines running for $n$ time points (for example the  layers  of the HMAX architecture in \cite{Riesenhuber1999}
can be   described as AND operations (dot products) followed by OR operations
(pooling), i.e. as  disjunctions of conjunctions.).
Given a very large set of labeled examples it is not too surprising
that greedy algorithms such as stochastic gradient descent can find
satisfactory parameters in such an architecture, as shown by the
recent successes of Deep Convolutional Networks. Supervised learning
with millions of examples, however, is not, in general, biologically
plausible. Our theory can be seen as proving that a form of
unsupervised learning in convolutional architectures is possible and
effective, because it provides invariant representations with small
sample complexity.\\
\noindent
{\it Two stages: group and non-group transformations.}
The core of the theory applies  to compact groups
such as rotations of the image in the image plane.  Exact invariance
for each module is equivalent to a localization condition
which could be interpreted as  a form of sparsity \cite{anselmi2013unsupervised}.
 If the condition is relaxed to hold approximately it becomes a {\it sparsity condition for the class of
  images w.r.t. the dictionary $t^k$ under the group $G$} when
restricted to a subclass of similar images.
This property, which is similar to compressive sensing ``incoherence'' (but in a group
context), requires that $I$ and $t^k$ have a representation with
rather sharply peaked autocorrelation (and correlation) and guarantees
approximate invariance for transformations which  do not have
group structure,  see \cite{leibo2014modularity}.\\
\noindent
{\it Robustness of pooling.} It is interesting that the theory is robust with respect to the
pooling nonlinearity.  Indeed, as discussed, very general class of nonlinearities will work, see Appendix A.
 Any nonlinearity will provide invariance, if the nonlinearity does not change with time and is the same for all the
simple cells pooled by the same complex cells.  A sufficient number of
different nonlinearities, each corresponding to a complex cell, can
provide  selectivity \cite{anselmi2013unsupervised}.\\
\noindent
{\it Biological predictions and biophysics, including dimensionality
  reduction and PCAs.} There are at least two possible biophysical models for the theory.
The first is the original Hubel and Wiesel model of simple cells
feeding into a complex cell. The theory proposes the "ideal"
computation of a CDF, in which case the nonlinearity at the output of
the simple cells is a threshold. A complex cell, summating the outputs
of a set of simple cells, would then represent a bin of the histogram;
a different complex cell in the same position pooling a set of similar
simple cells with a different threshold would represent another bin of
the histogram.\\
\noindent
The second biophysical model for the HW module that implements the
computation required by i-theory consists of a single cell where
dendritic branches play the role of simple cells (each branch
containing a set of synapses with weights providing, for instance,
Gabor-like tuning of the dendritic branch) with inputs from the LGN;
active properties of the dendritic membrane distal to the soma provide
separate threshold-like nonlinearities for each branch separately,
while the soma summates the contributions for all the branches. This
model would solve the puzzle that so far there seems to be no
morphological difference between pyramidal cells classified as simple
vs complex by physiologists.
Further if the synapses are Hebbian it can be proved that Hebb's rule, appropriately modified with a normalization factor, is an online algorithm to compute the eigenvectors of the input covariance matrix, therefore tuning the dendritic branches weights to principal components and thus providing an efficient dimensionality reduction.
\\
\noindent
{\it ($n\rightarrow 1$).}The present phase of Machine Learning is characterized by supervised learning algorithms relying on large
sets of labeled examples ($n\rightarrow\infty$). The next phase is likely to focus on algorithms capable of learning from very few labeled
examples ($n\rightarrow 1$), like humans seem able to do. We propose and analyze a possible  approach to this problem  based on the unsupervised, automatic learning of a good representation for supervised learning, characterized
by small sample complexity ($n$).  In this view we take a step towards
a  major challenge in learning theory beyond the supervised learning, that is
the problem of {\it representation learning}, formulated
here as the unsupervised learning of invariant representations that
significantly reduce the sample complexity of the supervised learning stage.

\section*{Acknowledgment}

We would like to thank the McGovern Institute
for Brain Research for their support. This research was
sponsored by grants from the National Science Foundation, AFSOR-THRL
(FA8650-05-C-7262). Additional support was provided by
the Eugene McDermott Foundation.



\begin{thebibliography}{99}

\bibitem{akhiezer1965classical}
\textsc{Akhiezer, N.}  (1965) \emph{The classical moment problem: and some
  related questions in analysis}, University mathematical monographs. Oliver \&
  Boyd.

\bibitem{AndoniBIW09}
\textsc{Andoni, A., Ba, K.~D., Indyk, P.  {\small \&} Woodruff, D.~P.}  (2009)
  Efficient Sketches for Earth-Mover Distance, with Applications.. in
  \emph{FOCS}, pp. 324--330. IEEE Computer Society.

\bibitem{anselmi2013unsupervised}
\textsc{Anselmi, F., Leibo, J.~Z., Rosasco, L., Mutch, J., Tacchetti, A.
  {\small \&} Poggio, T.}  (2013) Unsupervised Learning of Invariant
  Representations in Hierarchical Architectures. \emph{arXiv preprint
  1311.4158}.

\bibitem{AnselmiPoggio2014}
\textsc{Anselmi, F.  {\small \&} Poggio, T.}  (2010) {Representation Learning
  in Sensory Cortex: a theory}. \emph{CBMM memo n 26}.

\bibitem{MM2013}
\textsc{Anselmi F. Leibo J.Z. Rosasco L. Mutch~J., T. A. P.~T.}  (2013) {Magic
  Materials: a theory of deep hierarchical architectures for learning sensory
  representations}. \emph{CBCL paper}.

\bibitem{Bengio2009}
\textsc{Bengio, Y.}  (2009) Learning Deep Architectures for AI.
  \emph{Foundations and Trends in Machine Learning 2}.

\bibitem{Berlinet}
\textsc{Berlinet, A.  {\small \&} Thomas-Agnan, C.}  (2004) \emph{Reproducing
  kernel Hilbert spaces in probability and statistics}. Kluwer Academic,
  Boston.

\bibitem{CramerWold1936}
\textsc{Cramer, H.  {\small \&} Wold, H.}  (1936) Some theorems on distribution
  functions. \emph{J. London Math. Soc.}, \textbf{4}, 290--294.

\bibitem{CucSma02}
\textsc{Cucker, F.  {\small \&} Smale, S.}  (2002) On the mathematical
  foundations of learning. \emph{Bulletin of the American Mathematical
  Society}, (39), 1--49.

\bibitem{Edelsbrunner10}
\textsc{Edelsbrunner, H.  {\small \&} Harer, J.~L.}  (2010) \emph{Computational
  Topology, An Introduction}. American Mathematical Society.

\bibitem{Fukushima1980}
\textsc{Fukushima, K.}  (1980) {Neocognitron: A self-organizing neural network
  model for a mechanism of pattern recognition unaffected by shift in
  position}. \emph{Biological Cybernetics}, \textbf{36}(4), 193--202.

\bibitem{burkhardt2007}
\textsc{Haasdonk, B.  {\small \&} Burkhardt, H.}  (2007) Invariant Kernel
  Functions for Pattern Analysis and Machine Learning. \emph{Mach. Learn.},
  \textbf{68}(1), 35--61.

\bibitem{Hein2005}
\textsc{Hein, M.  {\small \&} Bousquet, O.}  (2005) Hilbertian Metrics and
  Positive Definite Kernels on Probability Measures. in \emph{AISTATS 2005},
  ed. by Z.~G. Cowell, R., pp. 136--143. Max-Planck-Gesellschaft.

\bibitem{Hubel1962}
\textsc{Hubel, D.  {\small \&} Wiesel, T.}  (1962) {Receptive fields, binocular
  interaction and functional architecture in the cat's visual cortex}.
  \emph{The Journal of Physiology}, \textbf{160}(1), 106.

\bibitem{Hubel1965}
\textsc{Hubel, D.  {\small \&} Wiesel, T.}  (1965) {Receptive fields and functional architecture in two
  nonstriate visual areas (18 and 19) of the cat}. \emph{Journal of
  Neurophysiology}, \textbf{28}(2), 229.

\bibitem{Hubel1968}
\textsc{Hubel, D.  {\small \&} Wiesel, T.}  (1968) {Receptive fields and functional architecture of
  monkey striate cortex}. \emph{The Journal of Physiology}, \textbf{195}(1),
  215.

\bibitem{Boman2009}
\textsc{Jan~Boman, F.~L.}  (2009) Support Theorems for the Radon Transform and
  Cramér-Wold Theorems. \emph{Journal of Theoretical Probability,}.

\bibitem{JL84}
\textsc{Johnson, W.B.~Lindenstrauss, J.}  (1984) Extensions of Lipschitz
  mappings into a Hilbert space. \emph{Contemporary Mathematics}, \textbf{26}.

\bibitem{Kazhdan2003}
\textsc{Kazhdan, M., Funkhouser, T.  {\small \&} Rusinkiewicz, S.}  (2003)
  Rotation Invariant Spherical Harmonic Representation of 3D Shape Descriptors.
  in \emph{Proceedings of the 2003 Eurographics/ACM SIGGRAPH Symposium on
  Geometry Processing}, SGP '03, pp. 156--164.

\bibitem{lecun1995convolutional}
\textsc{LeCun, Y.  {\small \&} Bengio, Y.}  (1995) {Convolutional networks for
  images, speech, and time series}. \emph{The handbook of brain theory and
  neural networks}, pp. 255--258.

\bibitem{leibo2014modularity}
\textsc{Leibo, J.~Z., Liao, Q., Anselmi, F.  {\small \&} Poggio, T.}  (2014)
  The invariance hypothesis implies domain-specific regions in visual cortex.
  \emph{http://dx.doi.org/10.1101/004473}.

\bibitem{MallatGroupInvariantMain}
\textsc{Mallat, S.}  (2012) Group Invariant Scattering. \emph{Communications on
  Pure and Applied Mathematics}, \textbf{65}(10), 1331--1398.

\bibitem{Poggio2013}
\textsc{Poggio, T., Mutch, J., Anselmi, F., Tacchetti, A., Rosasco, L.  {\small
  \&} Leibo, J.~Z.}  (2013) {Does invariant recognition predict tuning of
  neurons in sensory cortex?}. \emph{MIT-CSAIL-TR-2013-019, CBCL-313}.

\bibitem{Ramm98onthe}
\textsc{Ramm, A.~G.}  (1998) On the theory of reproducing kernel hilbert
  spaces. .

\bibitem{reed1978}
\textsc{Reed, M.  {\small \&} Simon, B.}  (1978) \emph{Methods of modern
  mathematical physics. {II}. , Fourier Analysis, Self-Adjointness}. Academic
  Press, London.

\bibitem{Riesenhuber1999}
\textsc{Riesenhuber, M.  {\small \&} Poggio, T.}  (1999) {Hierarchical models
  of object recognition in cortex}. \emph{Nature Neuroscience}, \textbf{2}(11),
  1019--1025.

\bibitem{sermanet-iclr-14}
\textsc{Sermanet, P., Eigen, D., Zhang, X., Mathieu, M., Fergus, R.  {\small
  \&} LeCun, Y.}  (2014) OverFeat: Integrated Recognition, Localization and
  Detection using Convolutional Networks. in \emph{International Conference on
  Learning Representations (ICLR2014)}. CBLS.

\bibitem{soatto2009actionable}
\textsc{Soatto, S.}  (2009) Actionable information in vision. in \emph{Computer
  Vision, 2009 IEEE 12th International Conference on}, pp. 2138--2145. IEEE.

\bibitem{SrGrFu10}
\textsc{Sriperumbudur, B.~K., Gretton, A., Fukumizu, K., Sch{\"{o}}lkopf, B.
  {\small \&} Lanckriet, G. R.~G.}  (2010) Hilbert Space Embeddings and Metrics
  on Probability Measures. \emph{Journal of Machine Learning Research},
  \textbf{11}, 1517--1561.

\bibitem{Tosic2011}
\textsc{Tosic, I.  {\small \&} Frossard, P.}  (2011) Dictionary learning:
  {W}hat is the right representation for my signal?. \emph{IEEE Signal
  Processing Magazine}, \textbf{28}(2), 27--38.

\bibitem{Vapnik1982}
\textsc{Vapnik, V.}  (1982) \emph{{Estimation of dependencies based on
  empirical data}}. Springer Verlag.

\bibitem{vedaldi11efficient}
\textsc{Vedaldi, A.  {\small \&} Zisserman, A.}  (2011) Efficient Additive
  Kernels via Explicit Feature Maps. \emph{Pattern Analysis and Machine
  Intellingence}, \textbf{34}(3).

\bibitem{Xing03}
\textsc{Xing, E.~P., Ng, A.~Y., Jordan, M.~I.  {\small \&} Russell, S.}  (2003)
  Distance Metric Learning, With Application To Clustering With
  Side-Information. in \emph{Advances in neural information processing
  systems}, pp. 505--512. MIT Press.

\end{thebibliography}
%


%

\appendix
\section{Representation Via Moments}\label{Sec:Mom}

In Section \ref{Sec:CDFRep} we have discussed the derivation of invariant selective representation
considering the CDFs of suitable one dimensional probability distributions.
As we commented in Remark~\ref{Rem:MomRep} alternative representations are possible, for example
by considering moments. Here we discuss this point of view in some more detail.

Recall that if $\xi:(\Omega, p)\to \R$ is a random variables with law $q\in {\mathcal P}(\RR)$, then the associated moment vector is given is given by
\be\label{CDF}
m_q^r= {\mathbb E}|\xi|^r =\int dq |\xi |^r, \quad r\in N.
\ee
In this case we have the following definitions and results.
\bd[Moments Vector Map]\label{MomRep}
Let  ${\mathcal M}(\RR)=\{h~|~h:\N \to \R\}$, and
$${\mathcal M}(\RR)^\T=\{ h~|~h:\T\to
{\mathcal M}(\RR)\}.$$
Define
$$
M: {\mathcal P}(\RR)^\T  \to {\mathcal M}
(\RR)^\T, \quad M(\overline{\mu})(t)=m_{{\mu}^t}
$$
for $\overline \mu \in {\mathcal P}(\RR)$ and where we let $\overline{\mu}(t)=\overline{\mu}^t$, for
all $t\in \T$.
\ed
The above mapping associates to each  one dimensional distribution the corresponding vector of  moments. Recall that this association uniquely determines the probability distribution if the so called Carleman's condition is satisfied:
$$
\sum_{r=1}^{\infty}m_{2r}^{-\frac{1}{2r}} = +\infty
$$
where $m_{r}$ is the set of moments of the distribution.\\
We can then define the following representation.
\bd[Moments Representation]
Let
$$
\Sg:\XX\to {\mathcal M}(\RR)^{\T}, \quad \Sg=M\circ R\circ P,
$$
with $M$,$P$ and $R$ as in Definitions~\ref{MomRep},~\ref{ProbRep},~\ref{RadRep}, respectively.
\ed
Then, the following result holds.
\bt
For all $\x\in \XX$ and $t\in \T$
$$
\Sg^t(\x)(r)=\int dg |\scal{\x}{gt}|^r, \quad r\in \N,
$$
where we let $\Sg(\x)(t)=\Sg^t(\x)$. Moreover, for all $\x,\x'\in \XX$
$$
\x\sim \x'~~ \Leftrightarrow ~~ \Sg(\x)=\Sg(\x').
$$
\et
\bp
$\Sg = M\circ R\circ P$ is a composition of a one to one map $R$, a map $P$ that is one to one w.r.t. the equivalence classes induced by the group of transformations $\G$ and a map $M$ that is one to one since Carleman's condition is satisfied. Indeed, we have,
$$
\sum_{r=1}^{\infty}\Big(\int\;dg \scal{\x}{gt}^{2r}\Big)^{-\frac{1}{2r}}\leq
\sum_{r=1}^{\infty}\Big(\int\;dg |\scal{\x}{gt}|\Big)^{-\frac{1}{2r}2r} =  \sum_{r=1}^{\infty}
\frac{1}{C} = +\infty
$$
where $C=\int\;dg |\scal{\x}{gt}|$.
Therefore $\Sg$ is one to one w.r.t. the equivalence classes i.e. $\x\sim \x'~~ \Leftrightarrow ~~ \Sg(\x)=\Sg(\x')$.
\ep

We add one remark regarding possible developments of the above result.
\br
Note that the above result essentially depends on the characterization of the moment problem of probability distributions on the real line.  In this view, it could be further developed to consider for example the {\em truncated} case when only a finite number of moments is considered  or the generalized moments problem, where  families of (nonlinear) continuous functions, more general than powers, are considered (see e.g. \cite{akhiezer1965classical}).
\er

\section{Kernels on probability distributions}\label{RKHSProb}
To consider multi-layers within the framework proposed in the paper
we need to embed probability spaces in Hilbert spaces. A natural way to do so is by considering appropriate positive definite (PD) kernels, that is symmetric functions $K:X\times X\to \R$ such that
$$
\sum_{i,j=1}^nK(\rho_i, \rho_j)\alpha_i\alpha_j\ge 0
$$
for all $\forall \rho_1, \dots, \rho_n \in X , \alpha_1, \dots, \alpha_n\in \R$ and where
$X$ is any set, e.g. $X=\R$ or $X={\mathcal P}(\R)$.
Indeed, PD kernels are known to define a unique reproducing kernel Hilbert space (RKHS) $\hh_K$ for which they correspond to reproducing kernels,
in the sense that if $\hh_K$ is the RKHS defined by $K$, then $K_x=K(x, \cdot)\in \hh_K$
for all $x\in X$ and
\be\label{repro}
\scal{f}{K_x}_K=f(x), \quad \forall f\in \hh_K, x\in X,
\ee
where $\scal{\cdot}{\cdot}_K$ is the inner product in $\hh_K$  (see for example \cite{Berlinet} for an introduction to RKHS).\\
 Many examples of kernels on distributions are known and have been studied. For example   \cite{Hein2005,vedaldi11efficient} discuss a variety of  kernels of the form
$$
K(\rho, \rho')=\int \int  d \gamma(x) \kappa(p_\rho(x), p_{\rho'}(x))
$$
 where $p_\rho, p_{\rho'}$ are the densities of the measures $\rho, \rho'$ with respect to
 a dominating measure $\gamma$ (which is assumed to exist) and
 $\kappa:\R_0^+\times \R_0^+\to \R$ is  a PD kernel.
 Recalling that  a PD kernel defines a pseudo-metric via  the equation
$$
d_K(\rho, \rho')^2=K(\rho, \rho)+K(\rho', \rho)-2K(\rho, \rho').
$$
  it is shown in  \cite{Hein2005,vedaldi11efficient} how different classic metric on probability distributions can be recovered  by suitable choices of the kernel $\kappa$. For example,
 $$\kappa(x,x')=  \sqrt{xx'},$$
corresponds to the Hellinger's
distance,
see  \cite{Hein2005,vedaldi11efficient}  for other examples.

A different approach    is based on defining kernels of the form
\be\label{MeanK}
K(\rho, \rho')=\int \int  d\rho(x)d\rho'(x') k(x,x'),
\ee
where    $k:\R\times \R\to \R$ is a PD kernel.
Using the reproducing property of $k$ we can write
$$
K(\rho, \rho')=\scal{\int d\rho(x') k_x}{\int  d\rho(x) k_{x'}}_k=\scal{\Phi(\rho)}{\Phi(\rho')}
$$
where $\Phi:{\mathcal P}(\R)\to \hh$ is the embedding $\Phi(x)=\int d\rho(x') k_x$
mapping each distribution in a corresponding kernel {\em mean}, see e.g. \cite{Berlinet}. Condition on the kernel $k$, hence on $K$, ensuring that  the corresponding function $d_K$ is a metric have been studied in detail, see  e.g. \cite{SrGrFu10}.\\


\end{document}